\documentclass[10pt,twocolumn,letterpaper]{article}

\usepackage[pagenumbers]{cvpr} 

\definecolor{cvprblue}{rgb}{0.21,0.49,0.74}
\usepackage[pagebackref,breaklinks,colorlinks,allcolors=cvprblue]{hyperref}
\usepackage{multirow}
\usepackage{pifont}
\newcommand{\cmark}{\ding{51}}  
\newcommand{\xmark}{\ding{55}}  

\def\paperID{24} 
\def\confName{CVPR}
\def\confYear{2025}

\title{Mix-QSAM: Mixed-Precision Quantization of the Segment Anything Model}

\author{Navin Ranjan \hspace{0.75cm} Andreas Savakis\\
Rochester Institute of Technology\\
Rochester, New York 14623, USA\\
}
\begin{document}
\maketitle
\begin{abstract}
The Segment Anything Model (SAM) is a popular vision foundation model, however, its high computational and memory demands make deployment on resource-constrained devices challenging. While Post-Training Quantization (PTQ) is a practical approach for reducing computational overhead, existing PTQ methods rely on fixed bit-width quantization,  leading to suboptimal accuracy and efficiency. To address this limitation, we propose Mix-QSAM, a mixed-precision PTQ framework for SAM. First, we introduce a layer-wise importance score, derived using Kullback–Leibler (KL) divergence, to quantify each layer’s contribution to the model’s output.
Second, we introduce cross-layer synergy, a novel metric based on causal mutual information, to capture dependencies between adjacent layers. This ensures that highly interdependent layers maintain similar bit-widths, preventing abrupt precision mismatches that degrade feature propagation and numerical stability. Using these metrics, we formulate an Integer Quadratic Programming (IQP) problem to determine optimal bit-width allocation under model size and bit-operation constraints, assigning higher precision to critical layers while minimizing bit width in less influential layers. 
Experimental results demonstrate that Mix-QSAM consistently outperforms existing PTQ methods on instance segmentation and object detection tasks, achieving up to 20\% higher average precision under 6-bit and 4-bit mixed-precision settings, while maintaining computational efficiency.

\end{abstract}    
\section{Introduction}
\label{sec:intro}

The Segment Anything Model (SAM)~\cite{kirillov2023segment} is a powerful prompt-guided vision foundation model with impressive zero-shot capabilities across various computer vision tasks, including image segmentation~\cite{SAM_ke2023segment, SAM2023segment_medical, SAM_wu2023medical, SAM_Seg_zhou2025darksam}, object detection~\cite{SAM_OD_cai2024crowd, SAM_OD_meeran2024sam}, tracking~\cite{SAM_Track_yang2023, SAM_Tracking_cheng2023}, and other downstream tasks~\cite{SAM_OA_1_liu2019perceptual, SAM_OA_2_shen2023anything, SAM_OA_3_yu2023inpaint}.  Despite its remarkable performance and broad range of applications, SAM’s high computational and memory requirements make deployment on resource-constrained and edge devices challenging. 
In particular, its vision transformer-based image encoder significantly contributes to this overhead.

To address this issue, various model compression techniques have been explored, including network pruning~\cite{he2018soft_prun, yu2022width_prun}, knowledge distillation~\cite{chen2022dear_KD, lin2022knowledge_KD, SAM_KD_kelenyi2024sam}, quantization~\cite{li2023repq, AdaLog_wu2025, Mix_QViT_ranjan2025, PTQ4SAM_lv2024, PQ_SAM_liu2024pq}, and compact architecture design~\cite{Mobile_SAM_zhang2023faster, Fast_SAM_zhao2023fast, Efficient_SAM_xiong2024, Tiny_SAM_shu2023}. Among these, post-training quantization (PTQ) has gained significant attention as it reduces model precision without modifying the architecture. PTQ converts network parameters (weights and activations) into low-bit representations without retraining and requires a small set of unlabeled samples for calibration, making it significantly faster and more practical than retraining-based approaches. 

While PTQ has been successfully applied to Vision Transformers (ViTs) \cite{PTQ_lin2022fq_vit, PTQ_liu2023noisyquant, PTQ_yuan2022ptq4vit, AdaLog_wu2025, PTQ_ding2022APQViTtowards}, its direct application to SAM presents unique challenges. Unlike ViTs designed for image classification, SAM performs per-pixel predictions across diverse zero-shot datasets, making it highly sensitive to low-bit quantization. This sensitivity arises due to asymmetric activation distributions and extreme outliers, which degrade quantization performance. For example, post-key-linear activations in SAM exhibit bimodal distributions with two peaks and a central void, expanding activation ranges and complicating quantization~\cite{PTQ4SAM_lv2024}. Additionally, high-magnitude outliers in specific channels require excessive adjustments, further increasing computational overhead~\cite{PQ_SAM_liu2024pq}.

To address these issues, recent works have introduced targeted quantization strategies. PTQ4SAM~\cite{PTQ4SAM_lv2024} proposes the Bimodal Integration (BIG) strategy, which normalizes bimodal distributions using channel-wise sign factors. PQ-SAM~\cite{PQ_SAM_liu2024pq} mitigates outlier distortions through the Grouped Activation Distribution Transformation (GADT) method, which reshapes activations via learnable shifting and scaling for improved quantization stability.

However, existing PTQ methods for ViTs and SAM apply fixed bit precision across all layers, assuming equal importance toward the model’s output. This forces critical layers to operate at the same precision as less significant ones, missing optimization opportunities. Mixed-precision quantization (MPQ) addresses this by assigning different bit-widths to layers, preserving accuracy in crucial layers while enhancing efficiency with lower precision elsewhere. Prior MPQ approaches employ search-based~\cite{liu2021post_Rank_Aware_PTQ, lou2020autoq, xiao2023patch-wise}, criterion-based~\cite{dong2019hawq, dong2020hawqv2}, and explainability-based~\cite{Mix_QViT_ranjan2025, LRP_QViT_ranjan2024} techniques. However, these methods often suffer from high computational overhead or are impractical for SAM. Hessian-based techniques~\cite{wang2019haq, dong2019hawq, dong2020hawqv2, yang2023global_nvit, CLADO_deng2023mixed} estimate layer importance but incur significant computational costs. Recently, Layer-wise Relevance Propagation (LRP)-based explainability methods~\cite{LRP_QViT_ranjan2024, Mix_QViT_ranjan2025} have shown promise in classification-based ViTs but remain unsuitable for SAM due to its multi-modal prompting and mask decoder, which disrupts gradient flow and relevance propagation. 

In this work, we propose a mixed-precision quantization strategy for  SAM, guided by two pre-computed metrics: layer importance score and cross-layer synergy. The layer importance score is inspired by causal mutual information, providing an interpretable measure of each layer's contribution to SAM’s final segmentation output. To compute layer importance, we perturb each layer by zeroing out its activations and measure the resulting shift in the output distribution 
between the original and perturbed models 
using Kullback-Leibler (KL) divergence. This 
methodology isolates the true contribution of each layer by eliminating confounding influences from other layers. A higher KL divergence indicates a layer’s greater importance to model performance, while a lower value suggests lesser impact. 

Cross-layer synergy measures the interdependence between neighboring quantized layers with respect to the model output. This metric enables the layers with strong synergy to have similar bit precision, preventing bottlenecks. To determine bit allocations based on these metrics, 
we formulate an Integer Quadratic Programming (IQP) problem optimizing bit-width allocation for each layer under model size and bit-operation constraints. This ensures efficient quantization while preserving segmentation accuracy. We adopt PTQ4SAM~\cite{PTQ4SAM_lv2024} as the fixed-bit quantization framework, and apply our mixed-precision bit allocations. Our main contributions are as follows:
\begin{itemize}
    \item To the best of our knowledge, we propose the first mixed-precision post-training quantization method for the Segment Anything Model, called Mix-QSAM.
    \item Our mixed-precision bit allocation strategy leverages layer importance scores and cross-layer synergy to optimize bit-width allocation. 
    Layer importance ensures higher precision for critical layers, while cross-layer synergy 
    maintains similar bit-widths for interdependent layers. We formulate an integer quadratic programming problem to optimize bit-width allocation under model size and bit-operation constraints.
    \item Mix-QSAM introduces a new method for quantifying layer importance using causal mutual information, computed by measuring the output distribution shift between the original and perturbed models via KL divergence.
    \item Mix-QSAM introduces cross-layer synergy, a novel metric based on causal mutual information, to quantify inter-layer dependencies in mixed-precision quantization. It ensures that highly synergistic layers maintain similar bit-widths, preventing abrupt precision shifts, reducing quantization errors, and enhancing feature propagation.
    \item Extensive experiments on SAM-Base, SAM-Large, and SAM-Huge demonstrate that Mix-QSAM outperforms existing methods, achieving up to 17\% higher average precision on instance segmentation
    and object detection under 6-bit and 4-bit mixed-precision settings.
\end{itemize}

\section{Related Work}
\label{sec:related_work}

\subsection{Segment Anything Model}
The Segment Anything Model, developed by Meta AI Research, is a prompt-driven vision foundation model pre-trained on the SA-1B dataset \cite{SAM_ke2023segment}.
SAM demonstrates a strong zero-shot generalization, and numerous studies have investigated its applications in various domains~\cite{SAM_ke2023segment, SAM_wu2023medical, SAM_KD_kelenyi2024sam, SAM_Track_yang2023, SAM_Tracking_cheng2023, SAM_OA_2_shen2023anything, SAM_OA_1_liu2019perceptual, SAM_OA_3_yu2023inpaint}. 
Various research efforts have explored enhancing SAM’s efficiency, resulting in the development of lightweight alternatives such as MobileSAM~\cite{Mobile_SAM_zhang2023faster}, FastSAM~\cite{Fast_SAM_zhao2023fast}, EfficientSAM~\cite{Efficient_SAM_xiong2024}, and TinySAM~\cite{Tiny_SAM_shu2023}, along with quantization-based methods like PTQ4SAM~\cite{PTQ4SAM_lv2024} and PQ-SAM~\cite{PQ_SAM_liu2024pq}.
MobileSAM~\cite{Mobile_SAM_zhang2023faster} improves efficiency by replacing SAM’s ViT-based encoder with a lightweight CNN and leveraging knowledge distillation to preserve segmentation performance. Similarly, FastSAM~\cite{Fast_SAM_zhao2023fast} replaces the transformer architecture with a lightweight CNN and an additional segmentation branch. While these architectural modifications reduce computational demands, they still require extensive training on the SA-1B dataset, making them resource-intensive. In contrast, PTQ4SAM~\cite{PTQ4SAM_lv2024} and PQ-SAM~\cite{PQ_SAM_liu2024pq} apply post-training quantization to reduce model parameters and incorporate modifications to mitigate quantization-induced accuracy loss. These approaches require only a small amount of unlabeled calibration data, making them more efficient than compact model redesigns. However, since these PTQ methods rely on fixed-precision quantization, they do not fully exploit the benefits of mixed-precision strategies, leading to a suboptimal balance between accuracy and efficiency.
\subsection{Vision Transformer Quantization}
Quantization reduces memory and computational burden by representing network parameters in low-bit formats. PTQ is a fast, practical, and widely used quantization technique that directly quantizes models without retraining, requiring only a small set of unlabeled samples for calibration. It is broadly categorized into statistical-based and learning-based approaches, and our method is adaptable to both. 

Statistical PTQ optimizes quantization parameters to minimize errors. 
PTQ4ViT~\cite{PTQ_yuan2022ptq4vit} reduces errors in Softmax and GELU activations using twin uniform quantization and introduces a Hessian-guided metric for scale selection. FQ-ViT~\cite{PTQ_lin2022fq_vit} mitigates inter-channel variations in LayerNorm via a Power-of-Two Factor and optimizes Softmax quantization with Log-Int-Softmax. RepQ-ViT~\cite{li2023repq} and AdaLog~\cite{AdaLog_wu2025} employ adaptive quantization, dynamically switching between channel-wise and layer-wise quantization for LayerNorm while using fixed~\cite{li2023repq} or adaptive~\cite{AdaLog_wu2025} logarithmic bases for Softmax quantization to improve precision and efficiency. 

Learning-based PTQ extends statistical PTQ by applying reconstruction techniques to further reduce quantization error~\cite{adaround, qdrop, brecq, liu2023pd, lee2023flexround}. Typically, reconstruction methods calibrate the quantized model using a small unlabeled dataset to minimize the reconstruction loss between quantized and full-precision outputs or intermediate activations. This calibration is done layer-by-layer or block-by-block, enabling precise adjustment of quantization parameters like scale and zero-point. AdaRound~\cite{adaround} optimizes weight rounding to minimize overall loss. BRECQ~\cite{brecq} introduces block-wise reconstruction to enhance quantization, while QDrop~\cite{qdrop} improves model robustness by introducing dropout operations during reconstruction.
 
However, uniform quantization, used by these methods, often yields suboptimal solutions. Layer importance score plays a crucial role in mixed-precision quantization by guiding bit-width allocation to balance accuracy and efficiency. Various methods have been explored to assess layer sensitivity to quantization, including Hessian-based techniques~\cite{dong2019hawq, dong2020hawqv2, CLADO_deng2023mixed, yang2023global_nvit}, gradient-based heuristics~\cite{gradient_li2024differentiable}, information-theoretic measures~\cite{IT_chen2024effective, MI_fan2021layer, MI_westphal2024mutual}, and learning-based strategies~\cite{luo2023deepburning}. Hessian-based methods estimate sensitivity using second-order loss curvature, where higher eigenvalues indicate greater quantization impact. However, their computational cost requires approximation techniques, such as diagonal Hessian estimation\cite{dong2020hawqv2} or Fisher information-based metrics\cite{fisher_shang2024quantized}. LRP-QViT\cite{LRP_QViT_ranjan2024} and Mix-QViT\cite{Mix_QViT_ranjan2025} leverage Layer-wise Relevance Propagation (LRP) explainability techniques to determine each layer’s contribution to the final prediction, enabling mixed-precision bit allocation based on relevance scores.

\section{Method}
\label{sec:method}
In this section, we present our Mix-QSAM framework in detail.
The Segment Anything Model consists of three main components: a ViT-based Image Encoder, a Prompt Encoder, and a Mask Decoder. The ViT encoder is the backbone of SAM, responsible for extracting high-resolution features but also being the primary source of computational complexity and model size. The Prompt Encoder processes user inputs (points, boxes, masks) into embeddings, while the Mask Decoder combines these embeddings with image features to generate segmentation masks.

\textbf{Vision Transformer.} The vision transformer reshapes an input image into a sequence of \(N\) flattened 2D patches, each mapped to a \(D\)-dimensional vector through a linear projection layer, denoted by $X\in{R}^{N\times D}$. The ViT consists of multiple transformer blocks, each with a multi-head self-attention (MHSA) module and a multi-layer perceptron (MLP) module. The MHSA module captures token relationships for global feature extraction. In the \(l^{th}\) block, the input \(X^l\) is projected to generate query, key, and value matrices: \(Q_l = X^lW^q_i\), \(K_l = X^lW^k_i\), and \(V_l = X^lW^v_i\), abbreviated as QKV. Attention scores are computed between the queries and keys, followed by a softmax operation, which is used to weight the values. The MHSA output is then generated by concatenating the weighted values from all attention heads, as follows:
\begin{align}
\text{Attn}_{i} = \text{Softmax}\left(\frac{Q_{i}\cdot K_{i}^T}{\sqrt{D_{h}}}\right)V_{i},
\label{eq:attn} \\
\text{MHSA}(X^{l}) = [\ \text{Attn}_{1}, \text{Attn}_{2},...,\text{Attn}_{i}]\ W^o,
\label{eq:attn_out}
\end{align}
where \(h\) is the number of attention heads, \(D_{h}\) is the feature size of each head, and \(i = 1, 2, \dots, h\). The MLP module consists of two fully connected layers separated by a GELU activation, projecting features into a high-dimensional space, as follows:
\begin{align}
\label{eq:mlp}
\text{MLP}(Y^{l}) = \text{GELU}\left(Y^{l}W^{1} + b^{1} \right)W^{2} + b^{2}.
\end{align}
Here, \(Y^{l}\) is the MLP input, \(W^{1} \in \mathbb{R}^{D \times D_f}\), \(b^{1} \in \mathbb{R}^{D_f}\), \(W^{2} \in \mathbb{R}^{D_f \times D}\), and \(b^{2} \in \mathbb{R}^{D}\). Layer normalization (\(LN\)) precedes each module, with residual connections added after. The transformer block is defined as:
\begin{align}
Y^{l} = X^{l} + \text{MHSA}\left(LN\left(X^{l}\right)\right),
\label{eq:mhsa_module} \\
X^{l+1} = Y^{l} + \text{MLP}\left(LN\left(Y^{l}\right)\right).
\label{eq:mlp_module}
\end{align}
The large matrix multiplications in MHSA and MLP significantly contribute to computational costs. To address this, we apply quantization to weights and activations across layers, including linear layers (QKV, Projection, FC1, FC2, and classifier head), the convolutional layer (Patch Embedding), and matrix multiplications (MatMul1, MatMul2), using a uniform quantizer. For power-law-like distributions, we adopt Adaptive Granularity Quantization (AGQ) from PTQ4SAM~\cite{PTQ4SAM_lv2024} for the post-Softmax layer. AGQ applies logarithmic quantization by searching for the optimal power-of-two base.

\subsection{Model Quantization}
The uniform quantization divides the data range into equal intervals and is defined as: 
\begin{align}
\text{Quant: } x^{q} = &\text{clip}\left(\left\lfloor\frac{x}{s}\right\rceil + z, 0, 2^{b}-1\right) \label{eq:uniform_quant} \\
\text{DeQuant: } \bar{x} &= s \cdot (x^{q} - z) \approx x \label{eq:uniform_qequant} \\
s = \frac{\max(x) - \min(x)}{2^b-1}, &\quad \text{and} \quad 
z = \left\lfloor -\frac{\min(x)}{s} \right\rceil \label{eq:scale_zeropoint}
\end{align}
Here, \(x\) represents the original floating-point weights or inputs, \(x^{q}\) is the quantized value, \(b\) is the quantization bit precision, \(\lfloor\cdot\rceil\) denotes rounding, and \(\text{clip}\) ensures values remain within the quantization range. The quantization scale \(s\) and zero-point \(z\) are determined by the lower and upper bounds of $x$. During dequantization,  \(\bar{x}\) approximately recovers the original \(x\).

Logarithmic quantization, particularly \(\log_2\), is a widely used approach for modeling power-law distributions. Recent works~\cite{li2023repq, AdaLog_wu2025, PTQ4SAM_lv2024} explores various logarithmic bases to boost model performance. The generalized form is:
\begin{align}
\text{Quant: } x^{q}&= \text{clip}\left( 
\left\lfloor -\log_{a} \frac{x}{s} \right\rceil, 0, 2^{b} - 1 \right) \notag\\
 &= \text{clip}\left(\left\lfloor -\frac{\log_2 \frac{x}{s}}{\log_2 a} \right\rceil, 0, 2^{b} - 1 \right) \label{eq:generalized_log_quant} \\
\text{DeQuant: } \bar{x} &= s \cdot a^{-x^{q}} \approx x \label{eq:generalized_log_dequant}
\end{align}
Standard \(\log_2\) quantization corresponds to \(a = 2\), while \(a = \sqrt{2}\) yields the \(\log_{\sqrt{2}}\) variant~\cite{li2023repq}. Adaptive approaches~\cite{AdaLog_wu2025, PTQ4SAM_lv2024} further enhance performance by learning \(a\) during training.


\subsection{Causal Layer Importance}
Understanding the contribution of individual layers in deep neural networks is crucial for effective model quantization. To quantify layer importance, we introduce the layer-wise causal information score, 
inspired by
Mutual Information (MI). By perturbing a layer through activation zeroing and analyzing the resulting shift in the output distribution, we estimate its true causal influence on the model’s predictions.

\textbf{Mutual Information} quantifies the association between two variables: the model output (\( Y \)) and the activations of a layer (\( L_i \)). It measures how much knowing one reduces the uncertainty of the other. The mutual information between them, \(I(L_i; Y)\), is defined as:
\begin{equation}
    I(L_i; Y) = H(Y) - H(Y | L_i),
\end{equation}
where \( H(Y) \) is the entropy of the output, and \( H(Y | L_i) \) is the conditional entropy given the layer’s activations. However, MI alone fails to distinguish causation from correlation, potentially overestimating a layer’s importance if it shares strong correlations with the output, even without directly influencing model decisions.

To address this, we introduce \textbf{Causal Mutual Information (CMI)}, which measures the direct impact of a layer by evaluating information loss when its activations are zeroed out. This isolates the true contribution of each layer by removing confounding influences from other layers \( L_{\neg i} \).  
\begin{equation}
    I_{\text{C}}(L_i; Y) = H(Y \mid L_{\neg i}) - H(Y \mid L_i, L_{\neg i}),
\end{equation}
where \( H(Y|L_{\neg i}) \) is the entropy when \( L_i \) is removed, and \( H(Y|L_i, L_{\neg i}) \) is the entropy when all layers are considered.

Direct entropy computation in high-dimensional spaces is intractable. While common methods such as histograms, Kernel Density Estimation (KDE), and K-Nearest Neighbors (KNN) exist, they suffer from either estimation errors or excessive high complexity. Instead, we adopt KL divergence to measure the shift in output distributions when a layer's activations are zeroed out:
\begin{align}
    I_{\text{C}}(L_i; Y) &\approx D_{\text{KL}}\left(P_Y(y \mid L_i, L_{\neg i}) \parallel P_Y(y \mid L_{\neg i})\right) \\
    = \sum_{y} &P_Y(y \mid L_i, L_{\neg i}) \log \frac{P_Y(y \mid L_i, L_{\neg i})}{P_Y(y \mid L_{\neg i})} \label{eq:CMI_Score}
\end{align}
where \( D_{\text{KL}} \) is the KL divergence, \( P_Y(y) \) is the marginal distribution of \( Y \), \( P_Y(y \mid L_i, L_{\neg i}) \) is the conditional distribution given \( L_i \) and \( L_{\neg i} \), and \( P_Y(y \mid L_{\neg i}) \) is the conditional distribution without \( L_i \). The summation \( \sum_{y} \) sums over all possible values of \( Y \) in the discrete case. A higher KL divergence indicates a greater influence of the layer on the model output, and vice versa.

To ensure comparability across layers, we normalize CMI scores by dividing each layer's score by the total CMI across all layers for a given image:
\begin{equation}
    \tilde{I}_{\text{C}}(L_i; Y) = \frac{I_{\text{C}}(L_i; Y)}{\sum_{l=1}^{L} I_{\text{C}}(L_l; Y)},
\end{equation}
where \( \tilde{I}_{\text{C}}(L_i; Y) \) represents the relative contribution of layer \( i \) to the model’s prediction, and \( L \) is the total number of layers in the network. Since CMI values vary across different inputs, we compute a generalized importance score (\(\Omega_i\)) by averaging the normalized CMI over \( T=64\) images:
\begin{equation}
    \Omega_i = \frac{1}{T} \sum_{t=1}^{T} \tilde{I}_{\text{C}}(L_i; Y_t).
\end{equation}
Here, \( Y_t \) represents the model output for the \( t \)-th image, and \( \Omega_i \) denotes the final importance score of layer \( i \), averaged over multiple images for generalization.

\begin{figure*}[thb]
\centering
\includegraphics[width=0.95\linewidth]{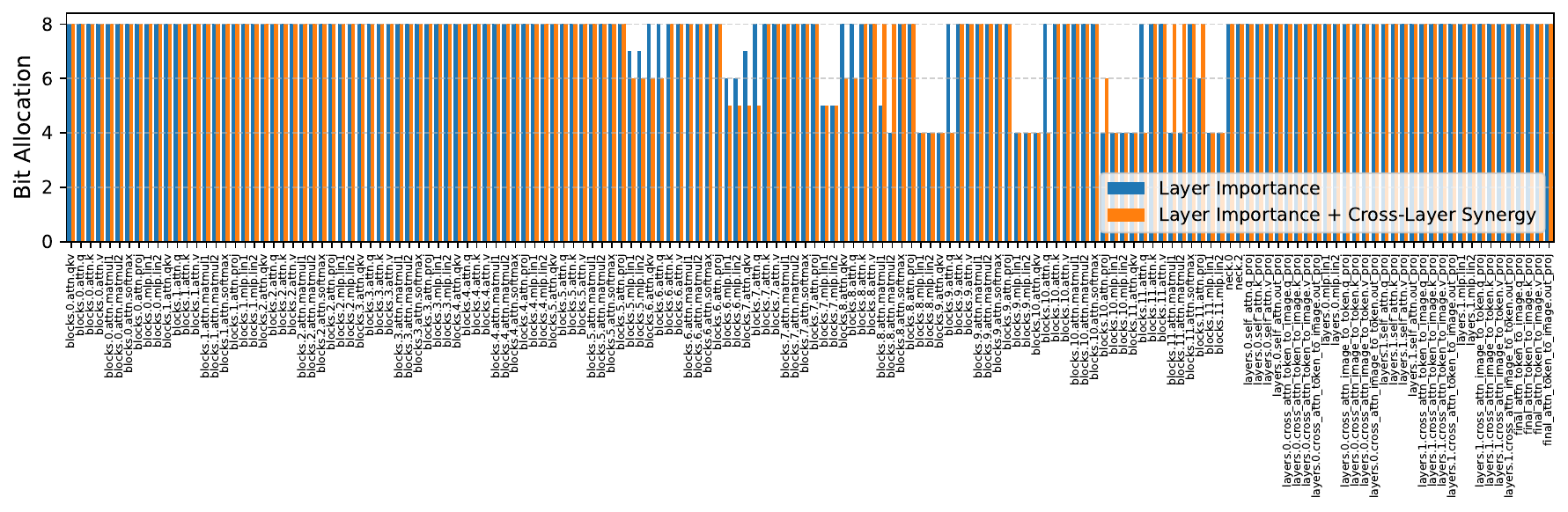}
  \caption{Mixed-precision bit allocation for SAM- under W6/A6: comparing layer importance score alone vs. using both layer importance score and cross-layer synergy. Candidate bit widths are \( \{4, 5, 6, 7, 8\}\)}
  \label{fig: Bit allocation comparision. }
\end{figure*}

\subsection{Cross-Layer Synergy}
Existing mixed-precision quantization methods~\cite{Mix_QViT_ranjan2025, LRP_QViT_ranjan2024, dong2019hawq, dong2020hawqv2, xiao2023patch-wise, SQNR_tai2024mptq}, typically allocate bit precision independently for each layer based on predefined metrics such as layer sensitivity or Hessian-based importance scores. However, this layer-wise approach ignores dependencies between adjacent layers, often leading to abrupt precision shifts, disrupted feature propagation, numerical instability, and information bottlenecks that degrade overall model performance. 

To address these limitations, we introduce a Bit Transition Regularization (BTR) method that prevents abrupt changes in bit precision between consecutive layers. A straightforward approach might consider the difference in bit precision between adjacent layers (independent BTR) ignoring inter-layer dependencies. However, this method uniformly applies regularization regardless of actual cross-layer dependence to model output. 

Instead, we propose a cross-layer synergy-based quantization approach that explicitly considers inter-layer dependencies to guide holistic bit-widths allocations. Instead of treating layers independently, our approach identifies strongly interdependent neighboring layers and maintains similar bit precision levels between them. This ensures smoother precision transitions, reduces quantization errors, enhances numerical stability, and preserves feature expressiveness, thereby improving overall model efficiency and performance.

We define the cross-layer synergy \(\mathcal{S}_{l,m}\) between two neighboring layers (\( L_l, L_{m} \)) based on causal mutual information, as follows:
\begin{equation}
\mathcal{S}_{l,m} = I(L_{l}, L_{m} ; Y) - I(L_{l} ; Y) - I(L_{m} ; Y), 
\end{equation}
where, \(I(L_{l}, L_{m};Y)\) represents the joint causal mutual information, quantifying the combined contribution of layers \( L_l\) and \(L_{m}\) to model output \(Y\). The individual causal mutual information scores, \(I(L_{l};Y)\) and \(I(L_{m};Y)\), measures each layer's independent contribution toward \(Y\). 

The synergy metric quantifies the additional information provided by considering both layers together, beyond their individual contributions. Higher synergy indicates strong interdependencies, suggesting these layers should maintain similar bit precision to prevent bottlenecks. Conversely, lower synergy values indicates layers contribute more independently, allowing more flexibility in bit allocation.

However, computing the full joint causal mutual information between adjacent layers significantly increases computational complexity. To mitigate this, we propose an efficient approximation based on the similarity in layer-to-output causal mutual information scores.
The cross-layer synergy is given as follows: 
\begin{align}
\mathcal{S}_{l,m}^t \approx &\frac{1}{(|I(L_{l} ; Y_t) - I(L_{m}; Y_t)| + \epsilon)} \\
\bar{\mathcal{S}}_{l,m} &= \frac{1}{T} \sum_{t=1}^{T} \mathcal{S}_{l,m}^{t} \\
\hat{\mathcal{S}}_{l,m} &= \log(1 + \bar{\mathcal{S}}_{l,m})
\end{align}
where, \(|I(L_{m} ; Y_t) - I(L_{m}; Y_t)|\) quantifies difference in individual contributions to the output, \(\epsilon\) is a small constant preventing numerical instability when the differences approach zero. The averaged synergy score \(\bar{\mathcal{S}}_{l,m}\) is computed over \( T=64\) samples, and the logarithmic transformation in \(\hat{\mathcal{S}}_{l,m}\) stabilizes numerical behaviors when layers have nearly identical contributions, preserves relative differences. 

\subsection{Mixed-Precision Bit Allocation}
Recent studies on CNNs~\cite{dong2019hawq,dong2020hawqv2,CLADO_deng2023mixed} and ViTs~\cite{SQNR_tai2024mptq,SQNR_lee2022optimizing,Hessian_Mixed_languagexu2021mixed,learned_layer_tang2022mixed,Mix_QViT_ranjan2025,LRP_QViT_ranjan2024} have explored mixed-precision quantization to balance accuracy and computational efficiency. MPQ assigns different bit-widths $(b_l^{w}, b_l^{a})$, selected from a predefined set $\mathcal{B}$, to weights and activations of each layer $l \in L$. The full model configuration is represented as \( \mathcal{F} = \{(b_l^{w}, b_l^{a})\}_{l=1}^{L} \), with the search space \( \mathcal{A} \) comprising all possible configurations.  The goal is to identify an optimal configuration $\mathcal{F}^{*} \in \mathcal{A}$ that maximizes validation accuracy under resource constraints $\mathcal{C}$. However, optimizing MPQ is computationally intensive, typically requiring iterative quantization, calibration, and evaluation, with methods like AutoQ~\cite{lou2020autoq} employing exhaustive search strategies to find optimal configurations.
In this work, we propose a mixed-precision bit selection strategy leveraging precomputed metrics: layer importance score (\(\Omega\)) and cross-layer synergy \(\hat{\mathcal(S)}\), to efficiently guide bit-width allocation. The layer importance score quantifies each layer's contribution to the model output, ensuring more critical layers receive higher precision.
The cross-layer synergy metric measures the interdependencies between neighboring layers, encouraging similar precision assignments to layers with strong dependencies, thus preventing numerical instability from abrupt precision changes. 
The mixed-precision bit allocation differences between using layer importance score only and using both layer importance score and cross-layer synergy are illustrated in Figure 1.

We formulate bit-width allocation as a constrained optimization problem to maximize the following objective function $\Phi$, defined as the weighted sum of layer importance scores multiplied by their assigned bit-widths, penalized by a bit transition regularizer to prevent abrupt precision changes between interdependent neighboring layers.
\begin{align}
&\Phi = \max_{\alpha} \Bigg( \sum_{l=1}^L  \Bigg( \Omega_l \odot \bigg( \sum_{j=1}^{|\mathcal{B}|} \alpha_{l, j} \cdot b_j \bigg)\Bigg) \notag \\ 
- \lambda & \Bigg(\sum_{l=1}^{L} \hat{\mathcal{S}}_{l,m} \cdot  \Bigg| \sum_{j=1}^{|\mathcal{B}|} \alpha_{l,j} \cdot b_j - \sum_{j=1}^{|\mathcal{B}|} \alpha_{m,j} \cdot b_j \Bigg| \Bigg)\Bigg), \label{eq:Mixed_precision_bit_allocation} \\
&\text{s.t.} \quad 
\alpha_{l, j} \in \{0, 1\}, \quad \sum_{j=1}^{|\mathcal{B}|} \alpha_{l, j} = 1, \quad \forall l, m, \label{eq:binary_constraint} \\
&\sum_{l=1}^L \sum_{j=1}^{|\mathcal{B}|} \alpha_{l, j} \cdot |w_l| \cdot b_j \leq \mathcal{C}_\text{M}, \label{eq:memory_constraint} \\
&\sum_{l=1}^L \sum_{j=1}^{|\mathcal{B}|} \text{MAC}_l \cdot \alpha_{l, j} \cdot b_j^2 \leq \mathcal{C}_{\text{BitOps}}, \label{eq:bitops_constraint} \\
&l \in \{1, \dots, L\}, \quad j \in \{1, \dots, |\mathcal{B}| \}. \label{eq:indices}
\end{align}
\noindent
Here, (\(l,m\)) denotes pairs of consecutive quantized  layers. Optimization constraints include target model size (\( \mathcal{C}_\text{M} \)) and bit operations (\( \mathcal{C}_{\text{BOP}} \)), ensuring efficient model deployment. For each quantized layer (\(l \)), we introduce one-hot selection variables \(\alpha_{l,j} \in \{0, 1\}^{|\mathcal{B}|}\), such that \(\sum_{j=1}^{|\mathcal{B}|} \alpha_{l, j} = 1\), ensuring each layer receives exactly one bit-width \(b_j\) from  \(\mathcal{B}\). Here, \(|w_l|\) represents the total number of parameters in layer \(l\), \( \text{MAC}_l \) denotes the number of multiply-accumulate (MAC) operations in layer \( l \), and \(\lambda \)=0.1 controls the impact of the smoothness penalty. This formulation provides a resource-aware mechanism, reducing computational overhead by eliminating iterative evaluations.

\section{Experiments}

\begin{table*}[t]
    \centering
    \small
    \begin{tabular}{llccccccc}
        \toprule
        \multirow{2}{*}{\textbf{Detector}} & \multirow{2}{*}{\textbf{Methods}} & 
        \multirow{2}{*}{\textbf{Rec.}} & \multicolumn{2}{c}{\textbf{SAM-B}} & \multicolumn{2}{c}{\textbf{SAM-L}} & \multicolumn{2}{c}{\textbf{SAM-H}} \\
        \cmidrule{4-9}
        &  & & W6/A6 & W4/A4  & W6/A6 & W4/A4 & W6/A6 & W4/A4 \\
        \midrule
        \multirow{7}{*}{Faster R-CNN \cite{ren2015faster}} 
        & Full-Precision&-&33.4&33.4&36.4&36.4&37.2&37.2\\
        & MinMax \cite{minmax}& \xmark &9.2 & - &32.9 & -  & 31.9 & - \\
        & Percentile \cite{percentile} & \xmark&10.9 & -  &33.5 & -  & 32.0 & -  \\
        & OMSE \cite{omse} &\xmark &11.9 & -  & 33.9 & 5.4   & 33.1 & 7.4  \\
        & PTQ4SAM-S~\cite{PTQ4SAM_lv2024} &\xmark&   15.4 & -   & 35.7 & 18.1   & 36.0&24.1  \\
        & AdaRound \cite{adaround}&\xmark& 23.1 &  - &34.3 & 8.7  & 33.7 & 14.5  \\
        & BRECQ \cite{brecq} &\cmark&24.1 &  - & 34.2 & 10.7   & 33.7 & 15.1  \\
        & QDrop \cite{qdrop} &\cmark& 29.3  & 13.0  &35.2 & 22.6   & 36.3 & 32.3  \\
        & PTQ4SAM-L~\cite{PTQ4SAM_lv2024}&\cmark & \underline{30.3} & \underline{16.0}   & \underline{35.8} & \underline{28.7}   & \underline{36.5} & \textbf{33.5}  \\
        & \textbf{Mix-QSAM(ours)}$^*$&\xmark   & \textbf{30.7} & \textbf{20.3}   & \textbf{35.9} & \textbf{28.8}   & \textbf{36.6}& \underline{32.8} \\
        \midrule
        \multirow{7}{*}{YOLOX \cite{yolox}} 
        & Full-Precision&-&37.0&37.0&40.4&40.4&41.0&41.0\\
        & MinMax \cite{minmax}&\xmark & 10.7 & -  & 37.5 & -  & 36.1 & - \\
        & Percentile \cite{percentile} &\xmark& 12.0 & -  & 38.0 & -   & 36.3 & -  \\
        & OMSE \cite{omse} &\xmark& 13.5 & -  & 38.4 & 6.1  & 37.5 & 7.8 \\
        & PTQ4SAM-S~\cite{PTQ4SAM_lv2024}& \xmark  & 17.4 & -  & 40.0 & 20.6  & 40.3&26.7  \\
        & AdaRound \cite{adaround} &\cmark& 26.4 & -  & 38.9 & 11.1 & 38.3 & 16.7  \\
        & BRECQ \cite{brecq} &\cmark& 26.1 & - & 38.9 & 12.0  & 38.3 & 16.3  \\
        & QDrop \cite{qdrop} &\cmark& 33.6 & 13.3  & 39.7 & 25.3  & 40.4 & 35.8  \\
        & PTQ4SAM-L~\cite{PTQ4SAM_lv2024} &\cmark &\underline{34.3} & \underline{18.4}   & \textbf{40.3} & \textbf{31.6}  & \textbf{40.7} & \textbf{37.6}  \\
        & \textbf{Mix-QSAM(ours)}$^*$ &\xmark&  \textbf{35.3} & \textbf{26.9}  & \textbf{40.3} & \underline{30.1}  & \textbf{40.7}& \underline{36.1} \\
        \midrule
        \multirow{7}{*}{H-Deformable-DETR \cite{hdeformabledetr}} 
        & Full-Precision&-&38.2&38.2&41.5&41.5&42.0&42.0\\
        & MinMax \cite{minmax} &\xmark& 10.9 & -  & 38.6 & -  & 37.3 & -  \\
        & Percentile \cite{percentile} &\xmark& 12.3 & -   & 39.0 & -   & 37.5 & -  \\
        & OMSE \cite{omse} &\xmark& 15.0 & -  & 39.6 & 6.2  & 38.6 & 7.7  \\
        & PTQ4SAM-S~\cite{PTQ4SAM_lv2024} &\xmark&  17.9 & - &  41.0 & 20.9 &  41.3 & 27.3 \\
        & AdaRound \cite{adaround} &\cmark& 27.2 & -  & 39.9 & 8.0 & 39.4 & 16.3  \\
        & BRECQ \cite{brecq} &\cmark& 27.9 & -  & 39.9 & 11.1  & 39.5 & 15.5  \\
        & QDrop \cite{qdrop} &\cmark& 34.3 & 13.2  & 40.5 & 25.8  & 41.4 & 36.5  \\
        & PTQ4SAM-L~\cite{PTQ4SAM_lv2024} &\cmark& \underline{35.1} & \underline{17.3}  & \textbf{41.2} & \textbf{32.1}  & \textbf{41.6} & \textbf{38.4}  \\
        & \textbf{Mix-QSAM(ours)}$^*$ &\xmark& \textbf{35.6} & \textbf{27.1}  & \textbf{41.2} & \underline{31.4} &  \underline{41.5}& \underline{37.3} \\
        \midrule
        \multirow{7}{*}{DINO \cite{dino}}
        & Full-Precision&-&44.5&44.5&48.6&48.6&49.1&49.1\\
        & MinMax \cite{minmax} &\xmark& 11.2 & -  & 44.7 & - & 42.8 & -  \\
        & Percentile \cite{percentile} &\xmark& 14.0 & - &  45.4 & - & 43.1 & -  \\
        & OMSE \cite{omse} &\xmark& 16.6 & -  & 45.9 & 6.8  & 44.5 & 8.3  \\
        & PTQ4SAM-S~\cite{PTQ4SAM_lv2024} &\xmark& 20.4 & - & 47.7 & 23.1 &  48.1 & 30.5 \\
        & AdaRound \cite{adaround} &\cmark& 31.2 & 1.2 & 46.6 & 8.8 & 46.0 & 18.2  \\
        & BRECQ \cite{brecq} &\cmark& 31.8 & 3.6  & 46.6 & 12.3  & 46.0 & 17.6  \\
        & QDrop \cite{qdrop} &\cmark& 38.9 & 11.2  & 47.5 & 27.5  & 48.3 & 41.7  \\
        & PTQ4SAM-L~\cite{PTQ4SAM_lv2024} &\cmark& \underline{40.4} & \underline{14.4}  & \textbf{48.3} & \textbf{36.6}  & \textbf{48.7} & \textbf{43.9}  \\
        & \textbf{Mix-QSAM(ours)}$^*$ &\xmark&  \textbf{40.9} & \textbf{30.6} & \textbf{48.3} & \underline{33.9}  & \underline{48.5}& \underline{41.9} \\
        \bottomrule
    \end{tabular}
    \caption{Quantization results for image segmentation on the COCO~\cite{COCO_lin2014microsoft} dataset. Values represent mean Average Precision (mAP). Here, ``Rec.'' indicates whether reconstruction-based learning quantization is applied (\cmark for learning-based methods and \xmark for static methods), W/A indicates the bit-widths of weights and activations, and $*$'' denotes models with equivalent size and bit operations to fixed-bit quantized models. Bold indicates the best performance, and underline indicates the second-best performance.}
    \label{tab:quantization_comparison}
\end{table*}
For the layer importance score, we adopt the backbone and pretrained weights from~\cite{kirillov2023segment}. The mixed-precision bit allocation constraint optimization is solved using the CVXPY package~\cite{diamond2016cvxpy}. We build upon the PTQ4SAM~\cite{PTQ4SAM_lv2024} fixed-bit quantization framework, incorporating our mixed-precision method to develop Mix-QSAM. To ensure fair comparison, our mixed-precision models match their fixed-precision counterparts in model size and bit-operations.

We evaluate Mix-QSAM on two vision tasks—instance segmentation and object detection—using the COCO~\cite{COCO_lin2014microsoft} dataset, with mean Average Precision (mAP) as the evaluation metric. For instance segmentation, predicted bounding boxes from the detector serve as prompts for SAM to generate precise binary masks. For object detection, final bounding boxes are derived from segmentation masks using the Mask2Box method.

We use CNN-based detectors, Faster R-CNN~\cite{ren2015faster} and YOLOX~\cite{yolox}, along with transformer-based detectors, H-Deformable-DETR~\cite{hdeformabledetr} and DINO~\cite{dino}, to prompt SAM. Model calibration is performed using 32 randomly sampled training images. Following standard quantization practices~\cite{liu2023pd, qdrop, PTQ4SAM_lv2024}, we apply per-channel asymmetric quantization for weights and per-tensor asymmetric quantization for activations. Additionally, we keep the first and last layers unquantized, following~\cite{PTQ4SAM_lv2024, qdrop, PTQ_yuan2022ptq4vit}. Learning-based methods uses blockwise reconstruction with 20000 iterations.
  



\subsection{Results on Instance Segmentation}
We compare our proposed Mix-QSAM with several existing quantization methods, including statistic-based methods such as MinMax~\cite{minmax}, Percentile~\cite{percentile}, OMSE~\cite{omse}, and PTQ4SAM-S~\cite{PTQ4SAM_lv2024}, as well as learning-based methods such as AdaRound~\cite{adaround}, BRECQ~\cite{brecq}, QDrop~\cite{qdrop}, and PTQ4SAM-L~\cite{PTQ4SAM_lv2024}. Detailed results are shown in Table~\ref{tab:quantization_comparison}. Our static mixed-precision method, Mix-QSAM, significantly outperforms existing state-of-the-art (SOTA) static methods. For example, at 4-bit precision, Mix-QSAM improves mAP from 20.6\% to 30.1\% on SAM-L and from 26.7\% to 36.1\% on SAM-H compared to PTQ4SAM-S method using the YOLOX detector. 

Furthermore, Mix-QSAM achieves comparable or superior performance relative to learning-based methods but requires only a fraction of their quantization time, as detailed in Table~\ref{tab:model_efficiency}. Specifically, 4-bit Mix-QSAM achieves 26.8\% vs. 18.4\% (SAM-B) and 30.1\% vs. 31.6\% (SAM-L) compared to PTQ4SAM-L using YOLOX. Similar performance improvements are observed across other detectors and SAM models. At 6-bit precision, our method achieves lossless performance without relying on reconstruction-based learning optimization. 

\begin{table}[t]
    \centering
    \resizebox{\columnwidth}{!}{%
    \begin{tabular}{ccccc}
        \hline
        \textbf{Model} & \textbf{Methods} & \textbf{Rec.} & \textbf{W6A6} & \textbf{W4A4} \\
        \hline
        \multirow{5}{*}{SAM-B} 
        & Full-Precision&-&48.0&48.0\\
        & PTQ4SAM-S~\cite{PTQ4SAM_lv2024} & \xmark & 44.5 & 42.3 \\
        & BRECQ~\cite{brecq} & \cmark & 45.8 & 44.1 \\
        & QDrop~\cite{qdrop} & \cmark & 46.0 & 44.2 \\
        & PTQ4SAM-L~\cite{PTQ4SAM_lv2024} & \cmark & 46.0 & 44.4 \\
        & \textbf{Mix-QSAM(ours)}$^*$ & \xmark & \textbf{47.8} & \textbf{45.2} \\
        \hline
        \multirow{5}{*}{SAM-L} 
        & Full-Precision&-&48.2&48.2\\
        & PTQ4SAM-S~\cite{PTQ4SAM_lv2024} & \xmark & 46.8 & 45.6 \\
        & BRECQ~\cite{brecq} &\cmark  & 47.9 & 46.5 \\
        & QDrop~\cite{qdrop} & \cmark & 48.0 & 47.7 \\
        & PTQ4SAM-L~\cite{PTQ4SAM_lv2024} & \cmark & \textbf{48.2} & \textbf{47.8} \\
        & \textbf{Mix-QSAM(ours)}$^*$ & \xmark & \textbf{48.2} & 47.7 \\
        \hline
        \multirow{5}{*}{SAM-H} 
        & Full-Precision&-&48.3&48.3\\
        & PTQ4SAM-S~\cite{PTQ4SAM_lv2024} & \xmark & 46.3 & 44.6 \\
        & BRECQ~\cite{brecq} & \cmark & 47.9 & 47.4 \\
        & QDrop~\cite{qdrop} & \cmark & 48.2 & 47.7 \\
        & PTQ4SAM-L~\cite{PTQ4SAM_lv2024} & \cmark & \textbf{48.3} & \textbf{47.9}\\
        & \textbf{Mix-QSAM(ours)}$^*$ & \xmark & \textbf{48.3} & \textbf{47.9} \\
        \hline
    \end{tabular}
    }
    \caption{Quantization results for object detection on COCO~\cite{COCO_lin2014microsoft} dataset. Each value represents mean Average Precision (mAP). Here, ``Rec.'' indicates reconstruction-based learning, W/A indicates the bit-widths of weights and activations.}
    \label{tab:object_detection_results}
\end{table} 
\subsection{Results on Object Detection}
We evaluate our Mix-QSAM for object detection on the COCO~\cite{COCO_lin2014microsoft} dataset, as shown in Table~\ref{tab:object_detection_results}. Our method achieves superior performance on SAM-B and delivers competitive results on SAM-L and SAM-H compared to both static and learning-based PTQ approaches. Notably, at 6-bit precision, Mix-QSAM improves mAP on SAM-B from 44.5\% (PTQ4SAM-S) and 46.0\% (PTQ4SAM-L) to 47.8\%. Similar improvements are observed across other SAM variants and bit-precision levels.
\subsection {Qualitative Results}
To evaluate the effectiveness of our method, we present qualitative instance segmentation results for 4-bit quantization of SAM on the COCO dataset, using YOLOX bounding boxes as prompts (Figure~\ref{fig:qualitative}). PTQ4SAM-S (without reconstruction) fails to segment objects accurately. Both PTQ4SAM-L (with reconstruction) and our Mix-QSAM preserve object integrity, produce clean boundaries, and retain fine-grained details. However, PTQ4SAM-L incurs significant computational overhead due to its blockwise reconstruction, which involves 20000 iterations. In contrast, Mix-QSAM delivers comparable visual quality without requiring reconstruction, making it a more efficient and practical solution for low-bit deployment.

\subsection{Ablation Study}
We conduct an ablation study to analyze the impact of the layer importance score ($\Omega$) and bit transition regularization on the mixed-precision optimization described in Eq.~(\ref{eq:Mixed_precision_bit_allocation}). Table~\ref{tab:ablation_study} presents the results for instance segmentation using the YOLOX~\cite{yolox} detector. 
By incorporating layer importance scores ($\Omega$), we obtain consistent performance gains compared to the baseline method. 
For instance, on SAM-B, using $\Omega$ improves W6/A6 performance from 17.4\% (PTQ4SAM-S) to 24.8\% (Mix-QViT) and achieves 18.3\% at W4/A4. Next, we evaluate BTR, considering two variations: Independent BTR ($\mathcal{I}$, without cross-layer synergy) and Synergistic BTR ($\hat{\mathcal{S}}$, with cross-layer synergy). Independent BTR with layer importance score significantly boosts performance, reaching 29.3\% (W6/A6) and 21.4\% (W4/A4) on SAM-B. Furthermore, cross-layer synergy-based BTR achieves the highest overall performance of 35.3\% (W6/A6) and 26.9\% (W4/A4) on SAM-B. Similar improvements are consistently observed on SAM-L, validating the benefits of layer importance scoring and synergy-based BTR.
\begin{table}[t]
    \centering
    \resizebox{\columnwidth}{!}{%
    \begin{tabular}{ccccccc}
        \toprule
        \multirow{2}{*}{\textbf{Model}} & \multirow{2}{*}{\textbf{Methods}} &\multirow{2}{*}{\textbf{ $\Omega$ }} & \multicolumn{2}{c}{\textbf{BTR}} & \multirow{2}{*}{\textbf{W6/A6}} & \multirow{2}{*}{\textbf{W4/A4}} \\
        \cmidrule(){4-5}
        &&& $\mathcal{I}$ & $\hat{\mathcal{S}}$ & & \\
        \midrule
        \multirow{5}{*}{SAM-B}
        &PTQ4SAM-S~\cite{PTQ4SAM_lv2024}&\xmark&\xmark&\xmark&17.4&-\\
         &Mix-QViT(ours)$^*$& \cmark  & \xmark & \xmark & 24.8 &18.3  \\
         &Mix-QViT(ours)$^*$& \cmark & \cmark & \xmark & 29.3 & 21.4 \\
          &\textbf{Mix-QViT(ours)}$^*$& \cmark & \xmark & \cmark & \textbf{35.3} & \textbf{26.9}  \\
          \midrule
          \multirow{5}{*}{SAM-L}
        &PTQ4SAM-S~\cite{PTQ4SAM_lv2024}&\xmark&\xmark&\xmark&40.0&20.6\\
         &Mix-QViT(ours)$^*$& \cmark  & \xmark & \xmark & 40.1 &24.3  \\
         &Mix-QViT(ours)$^*$& \cmark & \cmark & \xmark & 40.1 & 26.5 \\
          &\textbf{Mix-QViT(ours)}$^*$& \cmark & \xmark & \cmark & \textbf{40.3} & \textbf{30.1}  \\
          
        \bottomrule
    \end{tabular}
    }
    \caption{Ablation study on the impact of layer importance score ($\Omega$) and Bit Transition Regularization (BTR) in MPQ of SAM. Each value denotes the mAP for instance segmentation using the YOLOX detector. Here, $\mathcal{I}$ indicates independent BTR, and $\hat{\mathcal{S}}$ represents synergy-based BTR.}
    \label{tab:ablation_study}
\end{table}
\subsection{Efficiency Analysis}
We analyze the efficiency of Mix-QSAM compared to existing quantization methods, as shown in Table~\ref{tab:model_efficiency}. Mix-QSAM achieves comparable or superior quantization accuracy 
without requiring the extensive reconstruction time of learning-based methods like PTQ4SAM-L. 
Although its calibration time is slightly longer than that of PTQ4SAM-S, Mix-QSAM delivers significantly higher accuracy with the same model size.
\begin{table}[t]
    \centering
    \resizebox{\columnwidth}{!}{%
    \begin{tabular}{cccccc}
        \toprule
        \multirow{2}{*}{\textbf{Model}} & \multirow{2}{*}{\textbf{Methods}} &  \multirow{2}{*}{\textbf{W6A6}} & \multicolumn{1}{c}{\textbf{Cali.}} & 
        \multicolumn{1}{c}{\textbf{Rec.}}  & \multicolumn{1}{c}{\textbf{MS}}\\
        &&&(min)&(min) & (MB)\\
        \midrule
        \multirow{5}{*}{SAM-B} 
        & Full-Precision&37.0&-&-&357.6\\
        & PTQ4SAM-S~\cite{PTQ4SAM_lv2024} & 17.4 &11.7&-&81.4 \\
        & BRECQ~\cite{brecq} & 26.1 & 13.4&217.2&81.4 \\
        & PTQ4SAM-L~\cite{PTQ4SAM_lv2024} & 34.3 &11.7&409.3&81.4 \\
        & \textbf{Mix-QSAM(ours)}$^*$ &  \textbf{35.3} & 15.1&-&81.6\\
        \midrule
        \multirow{5}{*}{SAM-L} 
        & Full-Precision&40.4&-&-&1168.6\\
        & PTQ4SAM-S~\cite{PTQ4SAM_lv2024} & 40.0 &16.5&-&242.0 \\
        & BRECQ~\cite{brecq} & 38.9 & 17.8&404.9& 242.0 \\
        & PTQ4SAM-L~\cite{PTQ4SAM_lv2024} & 40.3 &16.7&566.1& 242.0\\
        & \textbf{Mix-QSAM(ours)}$^*$ &  \textbf{40.3} & 15.1&-&241.9\\
        \bottomrule
    \end{tabular}
    }
    \caption{Efficiency analysis of different methods, comparing calibration time (Cali., minutes), reconstruction time (Rec., minutes; learning-based methods), and model size (MS) in MB.}
    \label{tab:model_efficiency}
\end{table}

\begin{table*}[t]
    \centering
\renewcommand{\arraystretch}{0.5}
\setlength{\tabcolsep}{2pt}
\begin{tabular}{ccccc}
  \includegraphics[width=3.0cm]{segmentation_fig/original/000000000785.jpg} & \includegraphics[width=3.0cm]{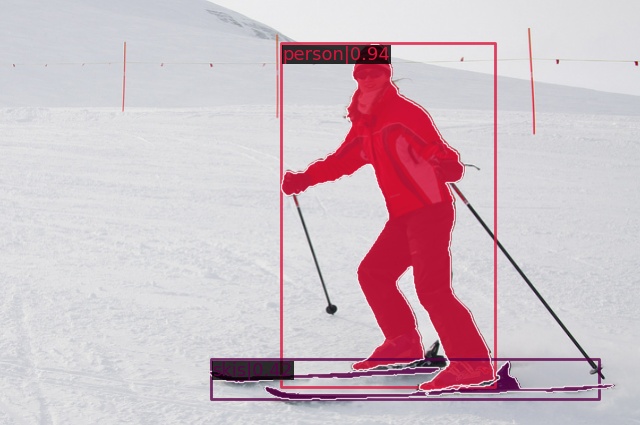} & \includegraphics[width=3.0cm]{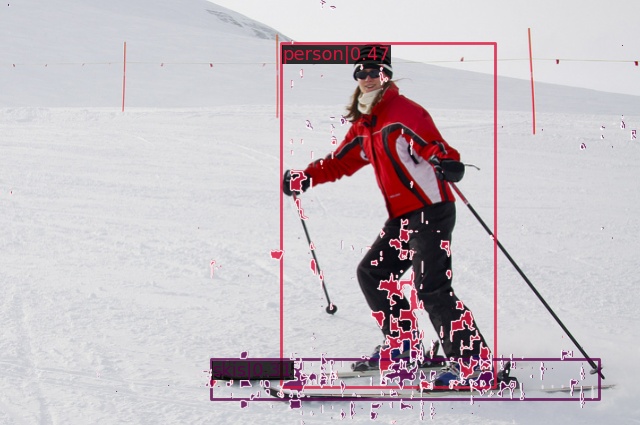} & \includegraphics[width=3.0cm]{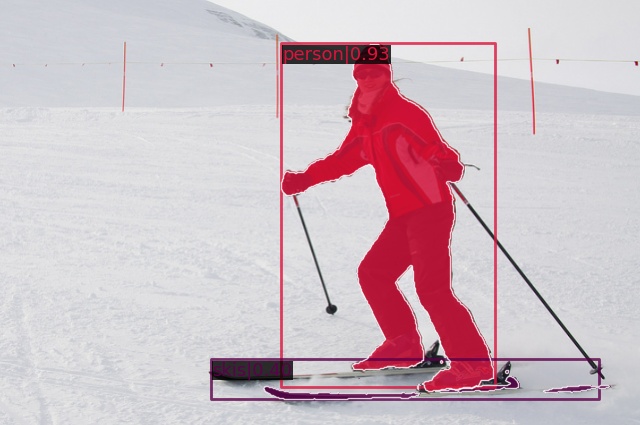} & \includegraphics[width=3.0cm]{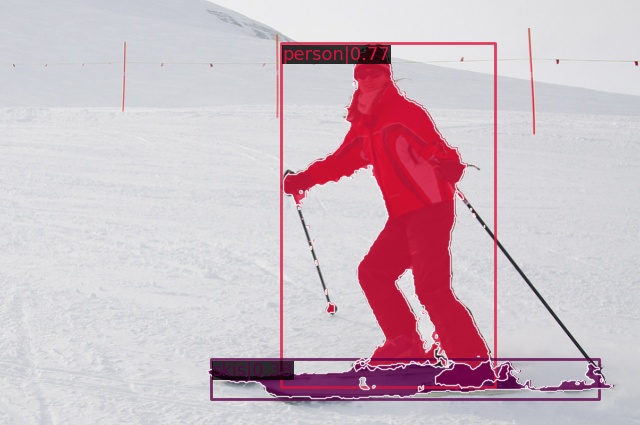} \\
  \includegraphics[width=3.0cm]{segmentation_fig/original/000000011149.jpg}  & \includegraphics[width=3.0cm]{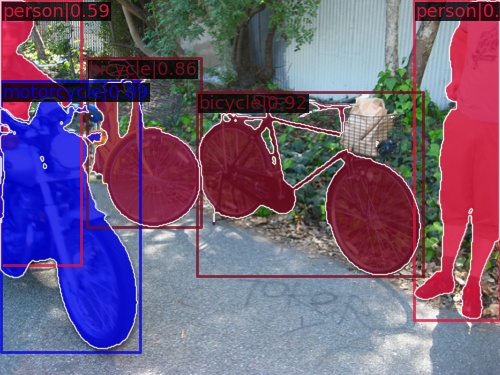}  & \includegraphics[width=3.0cm]{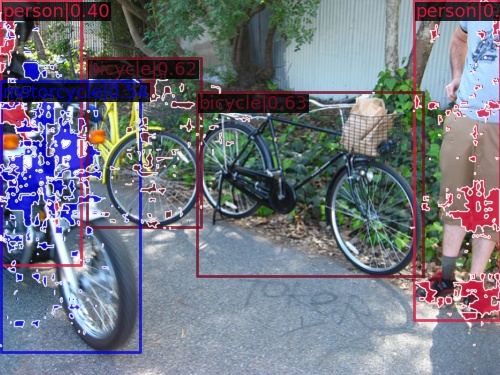}  & \includegraphics[width=3.0cm]{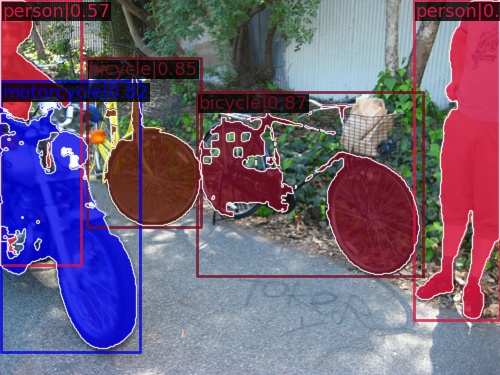}  & \includegraphics[width=3.0cm]{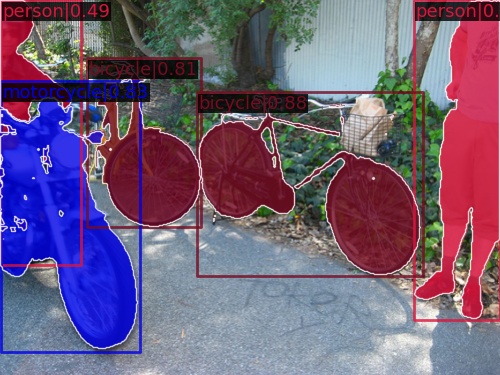} \\
  \includegraphics[width=3.0cm]{segmentation_fig/original/000000013291.jpg}  & \includegraphics[width=3.0cm]{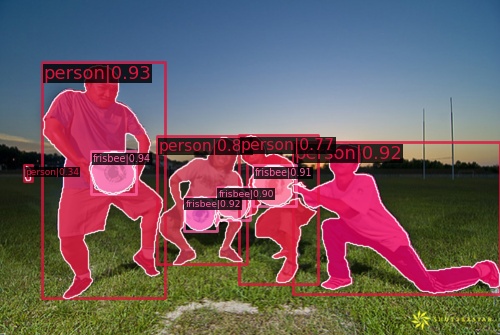}  & \includegraphics[width=3.0cm]{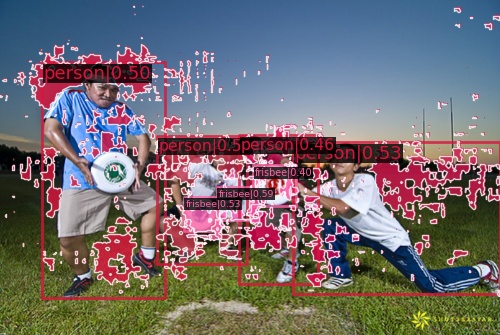}  & \includegraphics[width=3.0cm]{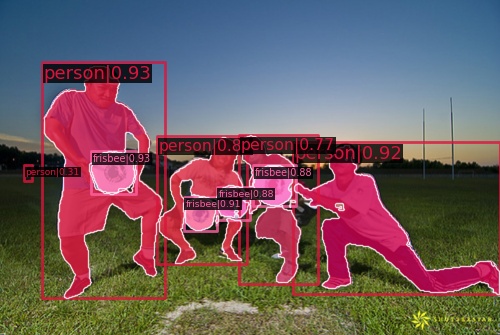}  & \includegraphics[width=3.0cm]{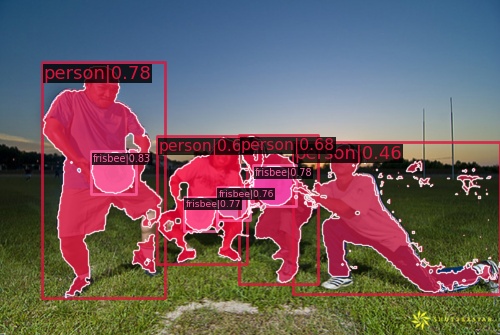} \\
  \includegraphics[width=3.0cm]{segmentation_fig/original/000000001532.jpg} & \includegraphics[width=3.0cm]{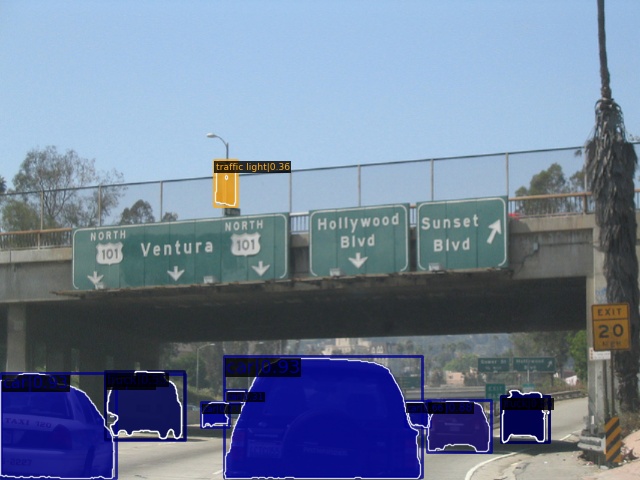} & \includegraphics[width=3.0cm]{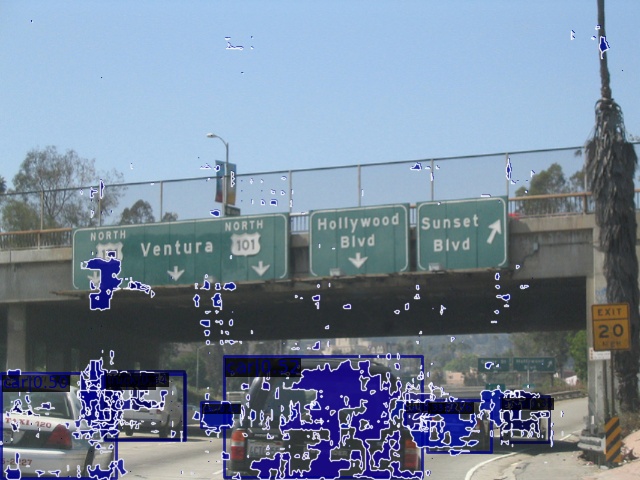} & \includegraphics[width=3.0cm]{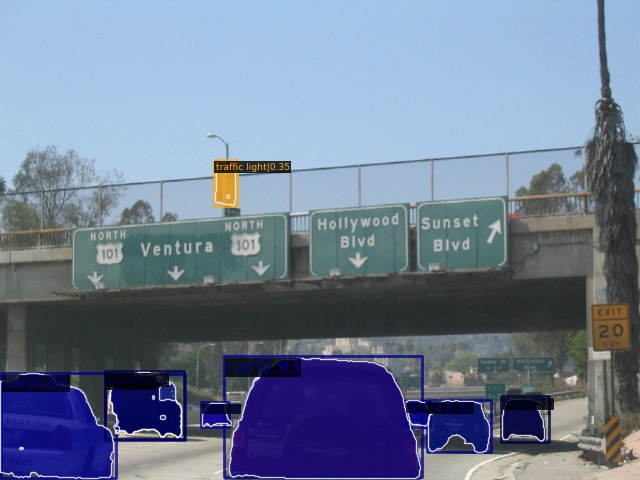} & \includegraphics[width=3.0cm]{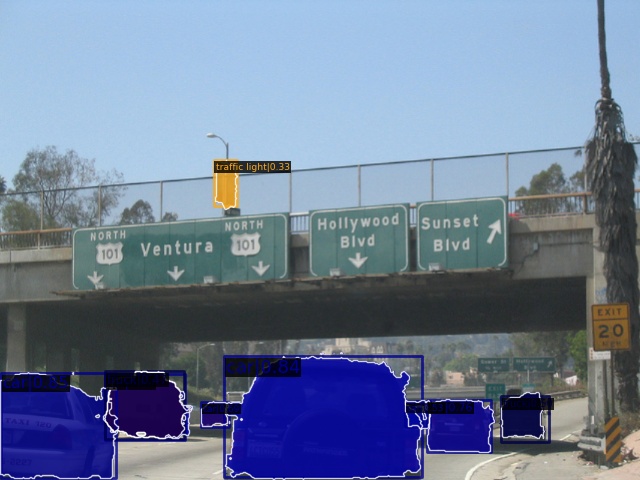} \\
  \includegraphics[width=3.0cm]{segmentation_fig/original/000000006040.jpg} & \includegraphics[width=3.0cm]{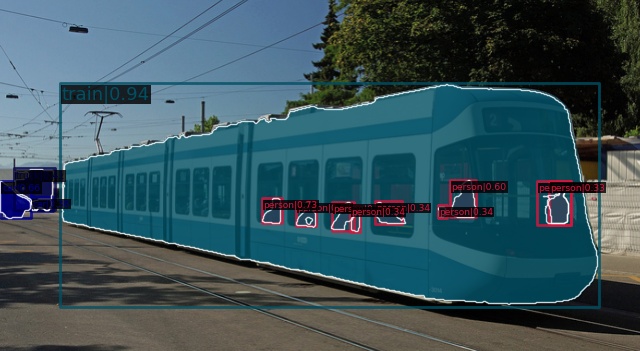} & \includegraphics[width=3.0cm]{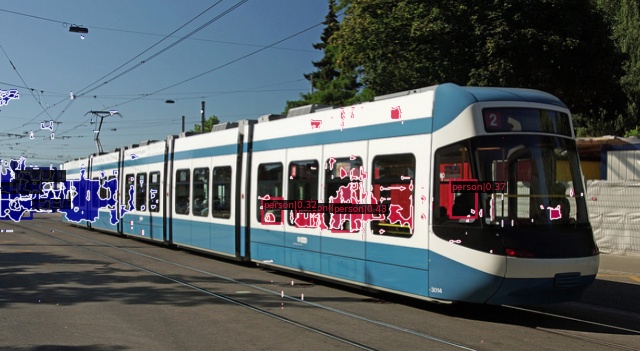} & \includegraphics[width=3.0cm]{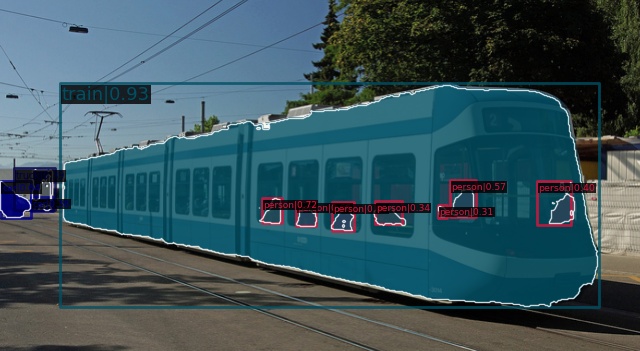} & \includegraphics[width=3.0cm]{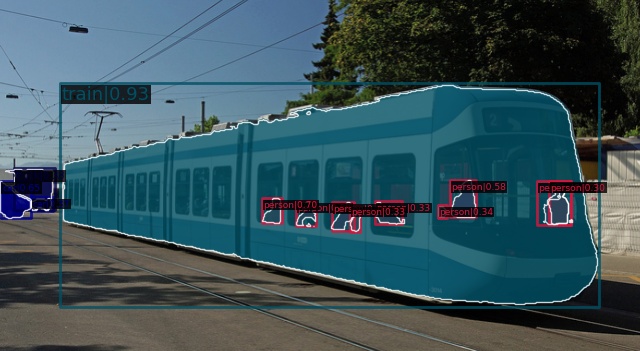} \\
  \includegraphics[width=3.0cm]{segmentation_fig/original/000000011760.jpg} & \includegraphics[width=3.0cm]{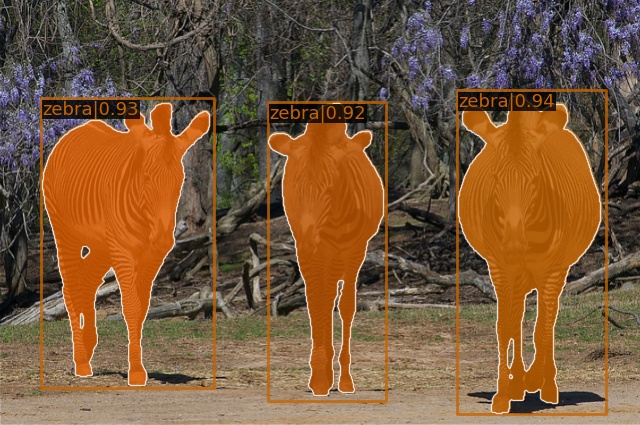} & \includegraphics[width=3.0cm]{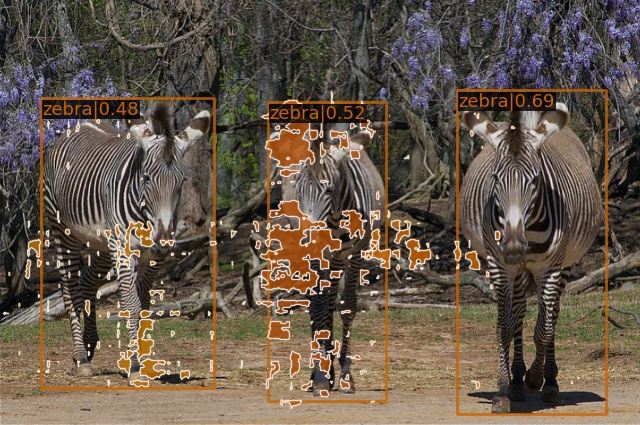} & \includegraphics[width=3.0cm]{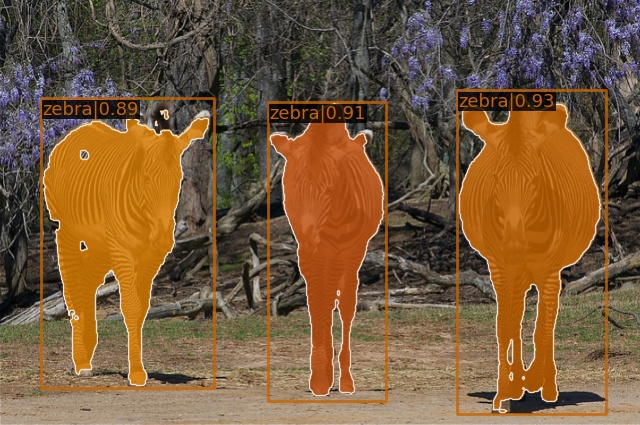} & \includegraphics[width=3.0cm]{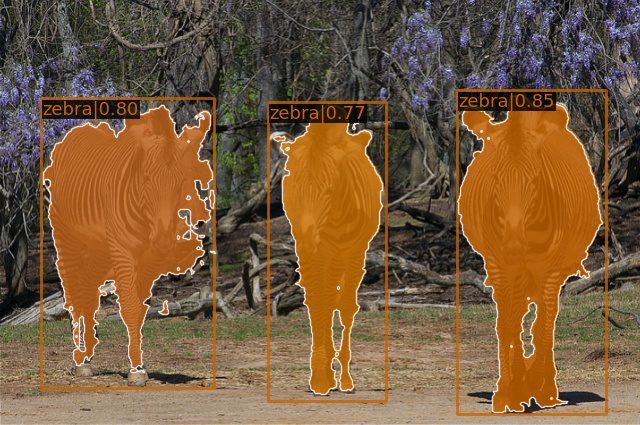} \\
  \makebox[3.0cm][c]{(a) Original} & 
  \makebox[3.0cm][c]{(b) Full Precision} & 
  \makebox[3.0cm][c]{(c) PTQ4SAM-S} & 
  \makebox[3.0cm][c]{(d) PTQ4SAM-L} & 
  \makebox[3.0cm][c]{(e) Mix-QSAM (Ours)} \\
\end{tabular}
\captionof{figure}{Qualitative comparison of instance segmentation on the COCO dataset using a 4-bit Segment Anything Model (SAM) with YOLOX bounding boxes as prompts. (a) Original images; (b) Full-precision SAM; (c) 4-bit PTQ4SAM-S without reconstruction; (d) 4-bit PTQ4SAM-L with reconstruction; (e) 4-bit Mix-QSAM (ours) without reconstruction.}
\label{fig:qualitative}
\vspace{1em}
\end{table*}

\section{Conclusion}
In this paper, we propose Mix-QSAM, a mixed-precision quantization framework for the Segment Anything Model, employing a KL-divergence-based layer importance score and a novel cross-layer synergy metric to guide bit-width allocation.
Experiments on instance segmentation and object detection demonstrate that our static PTQ-based Mix-QSAM achieves superior accuracy–efficiency trade-offs compared to existing learning-based PTQ methods.



\clearpage

\end{document}